\documentclass[review]{elsarticle}

\usepackage{hyperref}
\usepackage{amsmath}
\journal{Pattern Recognition}

\emergencystretch=\maxdimen
\begin{document}

\begin{frontmatter}

\title{Correlation-aware Adversarial Domain Adaptation and Generalization}

\author[1]{Mohammad Mahfujur Rahman \corref{cor1} }
\cortext[cor1]{Corresponding author: 
  Tel.: +61-416-453-032;  
  }
\ead{m27.rahman@qut.edu.au}
\author[1]{Clinton Fookes}
\author[2]{Mahsa Baktashmotlagh}
\author[1]{Sridha Sridharan}

\address[1]{Speech, Audio, Image and Video Technology (SAIVT) Laboratory - Queensland University of Technology - 2 George Street, Brisbane, QLD 4000, Australia}
\address[2]{University of Queensland
- Saint Lucia, QLD 4072, Australia}

\begin{abstract}
Domain adaptation (DA) and domain generalization (DG) have emerged as a solution to the domain shift problem where the distribution of the source and target data is different. The task of DG is more challenging than DA as the target data is totally unseen during the training phase in DG scenarios. The current state-of-the-art employs adversarial techniques, however, these are rarely considered for the DG problem. Furthermore, these approaches do not consider correlation alignment which has been proven highly beneficial for minimizing domain discrepancy. In this paper, we propose a correlation-aware adversarial DA and DG framework where the features of the source and target data are minimized using correlation alignment along with adversarial learning. Incorporating the correlation alignment module along with adversarial learning helps to achieve a more domain agnostic model due to the improved ability to reduce domain discrepancy with unlabeled target data more effectively. Experiments on benchmark datasets serve as evidence that our proposed method yields improved state-of-the-art performance.
\end{abstract}

\begin{keyword}
Domain adaptation, domain generalization, correlation-alignment, adversarial learning.
\end{keyword}
\end{frontmatter}

%\linenumbers

\section{Introduction}

\begin{figure}
\begin{center}
\includegraphics[width=1.0\linewidth]{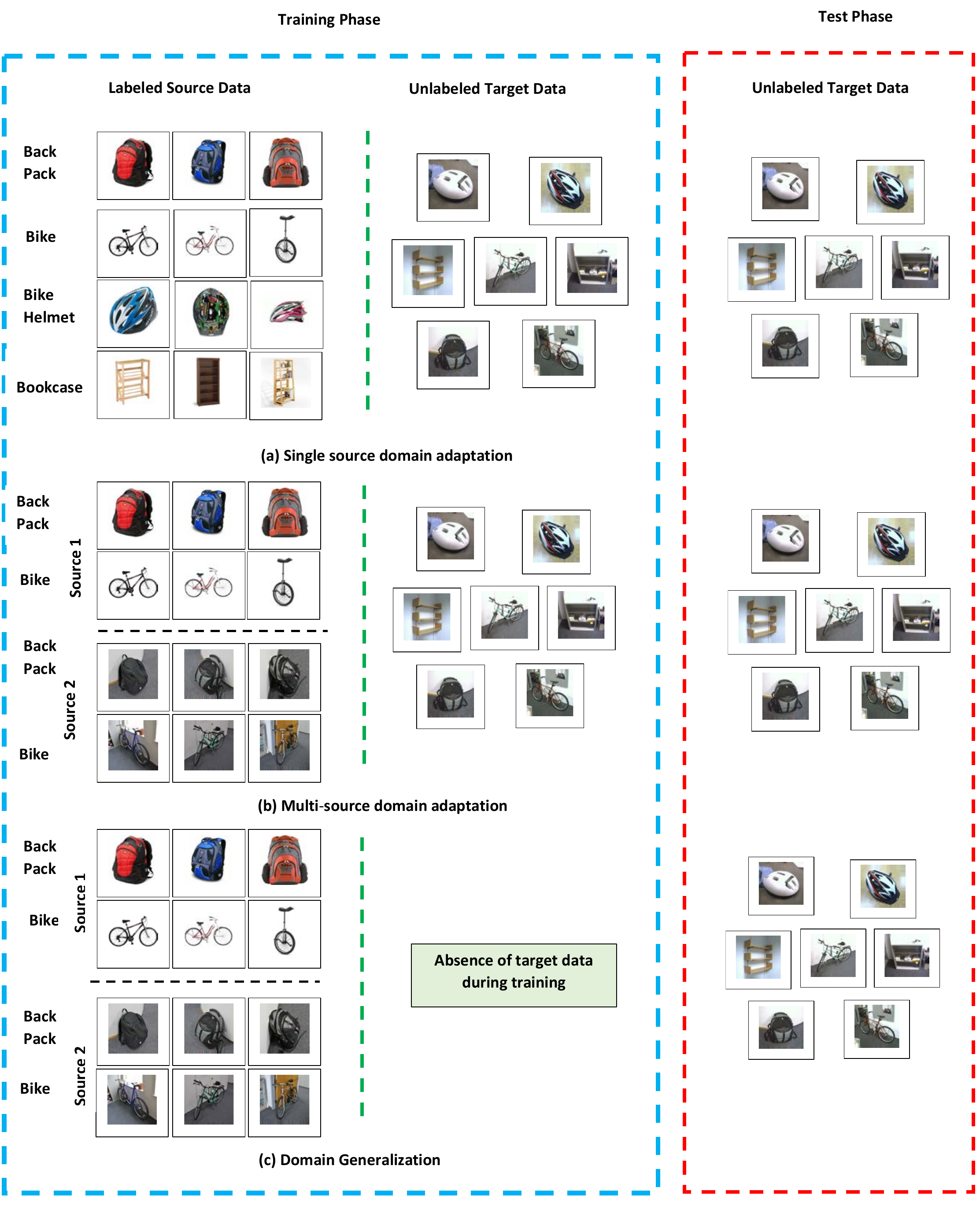}
\end{center}
   \caption{ (a) Single source domain adaptation can access the target data during training phase. (b) Multiple source domain adaptation can also access the target data during training phase. (c) Domain generalization cannot access the training data during training stage.}
\label{fig:DG_DA}
\end{figure}

In deep learning, it is commonly expected that the training and test data are collected from the same distribution and there will be a large set of labeled data. With this expectation, deep neural networks (DNNs) achieved tremendous success in various applications. Unfortunately, in many realistic scenarios, the above assumptions may not valid i.e., the training and test data may not have the same distribution and labeled data may not be available. Domain adaptation (DA) and domain generalization (DG) methods come forward to address this issue where previously labeled source data and a few or even no labeled target data is used to boost the task for a new domain. When the target data is unavailable during training, the typical approach is to utilize all the available source datasets. In the unavailability of the target data, DG techniques have been proposed which exploit all available source domain data that are less sensitive to the unknown target domain. DA methods require target data during training whereas DG methods do not require target data in the training phase. Figure \ref{fig:DG_DA} illustrates the difference between the DA and DG methods. DNNs \cite{7560644,6987333,7469327} allow us to extract powerful features  suitable for more types of input which are more domain invariant and transferable. Thus, researchers inspired to extend shallow DA techniques \cite{Ben-David2010,6751384,6909579,6247911,6751479} to deep DA techniques \cite{DBLP:journals/corr/TzengHZSD14,DBLP:conf/icml/LongC0J15,DBLP:conf/nips/LongZ0J16,DBLP:conf/icml/LongZ0J17,dcoral,7410820,pmlr-v37-ganin15,8099799,carlucci2017auto,NIPS2016_6254} and shallow DG techniques \cite{Khosla:2012:UDD:2402940.2402953,DBLP:conf/iccv/GhifaryKZB15} to deep DG techniques \cite{8237853,8053784}. 

%In \cite{7560644,6987333,7469327}, Deep learning-based network are used to extract powerful features suitable for more types of input.

Recently deep adversarial domain adaptation techniques \cite{pmlr-v37-ganin15,8099799} have achieved promising success in reducing the domain shift between the source and target data. Adversarial deep domain adaptation approaches are homologous to generative adversarial networks (GANs)\cite{NIPS2014_5423}. In these methods, a feature extractor is used to extract deep features and a domain classifier is trained to identify whether the data comes from a source domain or target domain. These methods reduce the domain disparity of the source and target data by employing a discriminator via a gradient reversal layer (GRL). 

%As these methods do not consider any semantic information, the discriminator is less capable to map the features from same category nearby and the features form different category far.

Distribution matching metric-based domain adaptation methods have also shown favourable results in reducing domain shift over DNNs. Maximum mean discrepancy (MMD), Central Moment Discrepancy (CMD) and Correlation alignment (CORAL) are the most used metrics in these domain adaptation methods \cite{DBLP:conf/icml/LongC0J15,DBLP:journals/corr/ZellingerGLNS17,dcoral}. These approaches use a Siamese network architecture (two streams) where one stream is used to extract features for the source data and another stream is used for the target data. The discrepancy metrics are used between the fully connected (fc) layers of the two streams of CNNs to reduce the domain discrepancy in the training stage.   

%Although the success of reducing domain shift is achieved by these domain adaptation methods, still these methods need to be developed for getting improved or desired performance under domain shift. 

In this paper, we propose a correlation-aware adversarial DA and DG framework where the correlation metric is used jointly with adversarial learning to minimize the domain disparity of the source and target data. Using both of these strategies concurrently further enforces the features of the same classes to be mapped nearby and the features of different classes to be far apart in DA scenarios. This also enforces the discrepancies among the source domains to be further reduced in DG scenarios. This approach significantly enhances the ability of prior adversarial adaptation approaches through our additional proposed domain discrepancy module. The proposed method was evaluated on five benchmark datasets (Office-31 \cite{Saenko:2010:AVC:1888089.1888106}, Office-Home  \cite{venkateswara2017Deep}, ImageCLEF-DA\footnote{http://imageclef.org/2014/adaptation}, Office-Caltech \cite{DBLP:conf/cvpr/GongSSG12} and PACS \cite{8237853}) on standard unsupervised DA and DG settings. Experiments prove that our proposed model for unsupervised DA and DG yields state-of-the-art results.

The contribution of this work is fourfold::
\begin{itemize}
%\color{blue}
\item We implement  a  novel  deep  domain adaptation and generalization framework that enables generalization on unknown target data.

\item The proposed deep domain adaptation architecture jointly adapts features using correlation alignment with adversarial learning.

\item The proposed framework can be used for both domain adaptation and domain generalization without changing the architecture of the model.

\item We report competitive  accuracy  compared  to  the  state-of-the-art  methods on five datasets and achieve the best average image classification accuracies.

\end{itemize}

\section{Related Work}

\textbf{Domain adaptation.} Recently, DNNs have been extensively adopted in explicitly reducing domain divergence by learning more transferable features. Deep learning based adaptation techniques can be divided into four categories: distribution matching based approaches, adversarial learning based approaches, generative adversarial learning based approaches and batch normalization based approaches. 

Maximum mean discrepancy (MMD) is one of the popular metrics for distribution matching based deep domain adaptation methods. MMD based DA methods \cite{DBLP:conf/icml/LongC0J15,DBLP:conf/nips/LongZ0J16,DBLP:conf/icml/LongZ0J17} measure the distance between the mean embeddings of the probability distributions of the source and target data in a reproducing kernel Hilbert space with a kernel trick. Another distribution matching metric is correlation alignment (CORAL) which aligns the covariances (second order statistics) between the source and target domains. CORAL based deep DA methods \cite{dcoral,DBLP:journals/corr/KoniuszTP16,morerio2018minimalentropy} have also shown promising performance in reducing the domain discrepancy of two domains. Central Moment Discrepancy (CMD) \cite{DBLP:journals/corr/ZellingerGLNS17} is also used in deep domain adaptation techniques where the domain divergence is reduced by matching the higher order central moments of the probability distributions of the source and target domains.

Nowadays, adversarial learning is promisingly used in domain adaptation. Most of the adversarial learning based domain adaptation methods \cite{pmlr-v37-ganin15,8099799} adopt the idea from GAN \cite{NIPS2014_5423}. In these methods, a discriminator is trained to classify the sampled feature comes from the source domain or target domain. On the other hand, the feature extractor is trained to fool the discriminator. All the adversarial learning based methods for domain adaptation discussed here apply adversarial losses in the embedding space.

GAN based domain adaptation methods \cite{NIPS2016_6544,Gen2Adapt,Hong_2018_CVPR,Hu_2018_CVPR} are also embraced in domain adaptation. Liu et al. proposed coupled generative adversarial networks (CoGAN) \cite{NIPS2016_6544} to learn a joint distribution of the source and target data where two classifiers are used for two domains and the classifiers are adapted so that the source classifier can classify the target samples correctly. Another GAN based domain adaptation method is proposed in \cite{Gen2Adapt} where the network produce source-like images from the source and target embeddings. GAN based domain adaptation methods perform adaptation in the pixel space, by contrast, adversarial based domain adaptation techniques perform adaptation in the feature space. 

Batch Normalization based domain adaptation methods \cite{DBLP:journals/corr/LiWSLH16,carlucci2017auto} decrease the domain discrepancy by aligning the source and target distributions to a canonical one. Li et al. \cite{DBLP:journals/corr/LiWSLH16} proposed an unsupervised domain adaptation method based on batch normalization where the statistics in all batch normalization layers are modulated across the network. In \cite{carlucci2017auto}, the source and target distributions are matched to a reference one via domain adaptation layers.

\textbf{Domain generalization.} DG is a less explored area compared to the DA problem. Blanchard et al. \cite{NIPS2011_4312} proposed a DG model where all the available source domains are aggregated and they learn a support vector machine classifier. In \cite{Muandet:2013:DGV:3042817.3042820}, the task of DG is achieved by minimizing the discrepancy among the source domains by invariant transformation. Khosla et al. \cite{Khosla:2012:UDD:2402940.2402953} proposed a DG approach where the weights of the classifier is adjusted during training and the classifier is applied on an unseen dataset. In \cite{DBLP:conf/iccv/GhifaryKZB15}, an auto-encoder is used to extract domain-invariant features from the available source domains by multi-task learning. These approaches\cite{NIPS2011_4312,Muandet:2013:DGV:3042817.3042820, Khosla:2012:UDD:2402940.2402953,DBLP:conf/iccv/GhifaryKZB15} use shallow models. Motiian et al. \cite{motiian2017CCSA} proposed a supervised deep generalization network where the source domains are aligned by using contrastive semantic alignment loss. Li et al. \cite{Li2018eccv} proposed a DG framework based on a conditional invariant adversarial network.

Our work is related to the approach proposed by Ganin et al. \cite{pmlr-v37-ganin15} for unsupervised domain adaptation. They used a common feature extractor network to extract the features of the source and target data. By contrast, we use two feature extractor networks for the corresponding two domains to achieve more domain specific features. In addition, we extend the target feature extractor network by adding a fully connected layer where the output vector of the layer is equal to the number of categories of the target data. We reduce the domain discrepancy using a domain discriminator via a GRL in the lower fully connected layer ($fc_{B}$) of a DNN along with a correlation alignment module which is used between the last fully connected layers ($fc_{8}$) of two streams of CNNs. We also extend our framework on DG scenarios. This approach makes our model to be unified which can be applied on both DA and DG settings. The method proposed by Tzeng et al. \cite{8099799} is also related to our work where two different feature extractors by unsharing the weights are used with an adversarial network. The difference between our work and \cite{8099799} is that they pre-train a source encoder convolutional neural network (CNN) on source data, and then perform adversarial learning. By contrast, we do not need to pre-train the source encoder CNN on the source data, and we use the correlation alignment metric in the last fully connected layer of the source and target streams of CNNs.

%Our model is related with \cite{dcoral} where CORAL is used to minimize the domain discrepancy, by contrast this method is not adversarial learning based method as ours. 

%-------------------------------------------------------------------------

\section{Proposed Method}

\begin{figure}
\begin{center}
\includegraphics[width=0.87\linewidth]{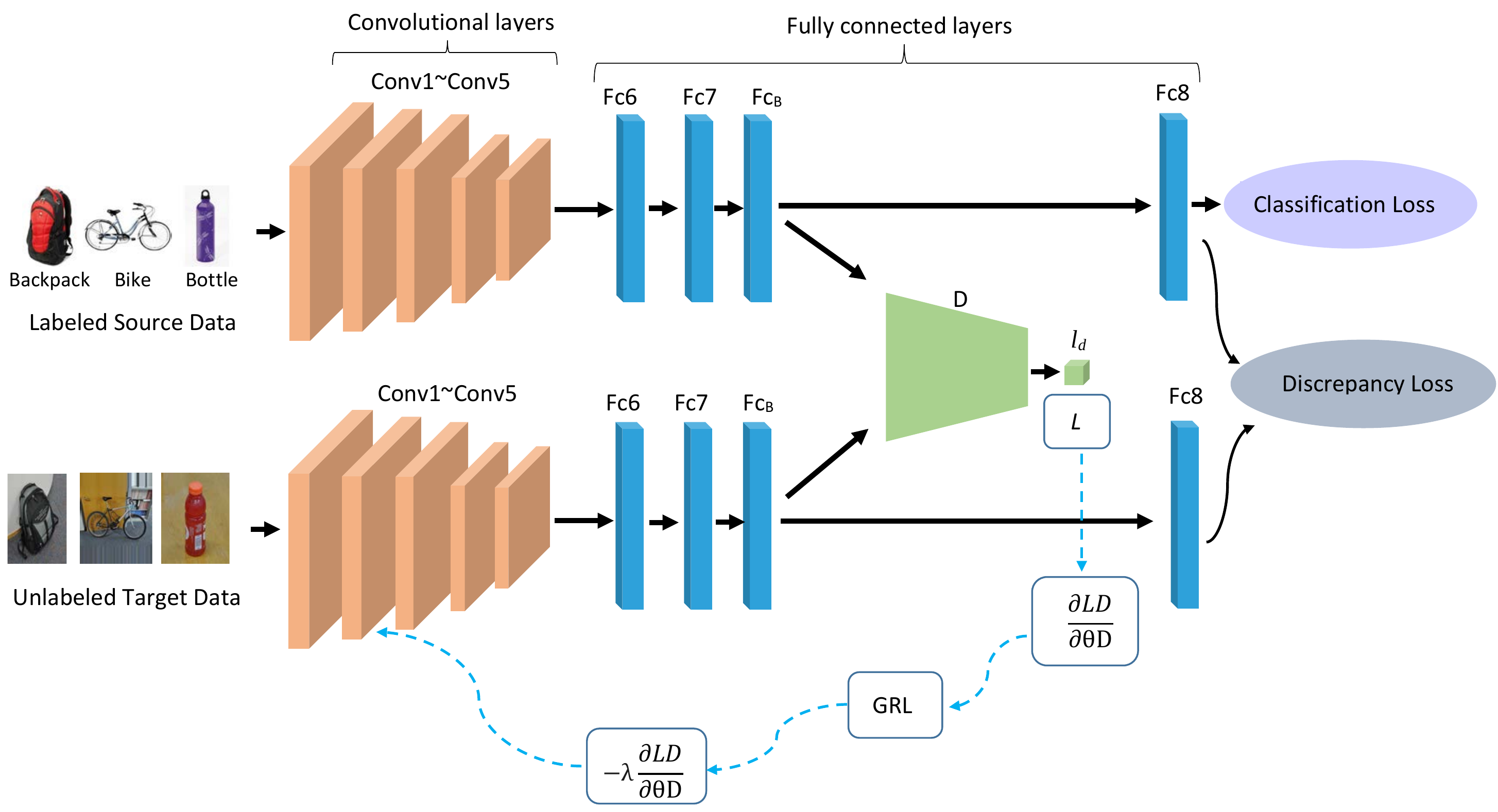}
\end{center}
   \caption{The proposed architecture for unsupervised domain adaptation. The domain discriminator, $D$, is used to minimize the domain discrepancy between the source and target domains along with the distribution matching metric (correlation alignment) which further enforces the alignment of the features of the source and target data nearby from the same classes and far for the features from different classes. The same architecture is used in DG scenarios where the discrepancy is minimized among the source domains.}
\label{fig:architecture}
\end{figure}

In this section, we present our proposed methodology for unsupervised domain adaptation and domain generalization in detail. We use the same network structure for domain adaptation and domain generalization methods. We consider the unsupervised domain adaptation issue where only labeled source data and unlabeled target data are available. The overview of our method is shown in Figure \ref{fig:architecture}. We use two feature extractor networks for obtaining more domain specific features and we extend the target feature network by integrating a fully connected layer where the output vector of this layer is equal to the number of categories of the target data. We use the correlation alignment metric between the last fully connected layers of the source and target feature extractor networks which enforce further to decrease the domain discrepancy of the source and target data. Before going into detail of our methodology for unsupervised DA and DG, we summarize the definitions of terminologies used.

It is assumed that there are $N_s$ number of source labeled data $\{X_i^s, Y_i^s\}$ and $N_t$ number of target data $\{X_i^t\}$ without their labels. It is also assumed that the data distribution of the source and target domain is different, i.e., $P_s(X_i^s,Y_i^s)$ $\neq$ $P_t(X_i^t,Y_i^t)$ where $Y_i^t$ is the label of target samples. The job of unsupervised domain adaptation is to gain a classifying model $F: X \rightarrow Y$ which is able to classify $\{X_i^t\}$ to the corresponding labels $\{Y_i^t\}$ given $\{X_i^s, Y_i^s\}$ and $\{X_i^t\}$ during training as the input. On the other hand, the job of domain generalization is to gain a classifying model $F: X \rightarrow Y$ which is able to classify $\{X_i^t\}$ to the corresponding labels $\{Y_i^t\}$ given $\{X_i^{s1}, Y_i^{s1}\}$, $\{X_i^{s2}, Y_i^{s2}\}$ $\dots$ $\{X_i^{sn}, Y_i^{sn}\}$ during training as the input and  $\{X_i^t\}$ is unavailable in the training stage where  $\{X_i^{s1}, Y_i^{s2}\}$, $\{X_i^{s2}, Y_i^{s2}\}$ and $\{X_i^{sn}, Y_i^{sn}\}$ are the samples from source domains $1$, $2$ and $n$ respectively.

\subsection{Adversarial Domain Adaptation}

A domain classifier is applied in a general feed-forward framework to form an adversarial domain adaptation model \cite{pmlr-v37-ganin15,8099799}. The general idea of this model is to learn domain invariant and class discriminative features. These models are analogous to the GAN with min-max loss of $F_s$ and $F_t$, the feature extractors for source and target domains respectively, and $D$, a binary classifier that can classify whether the features come from the source domain or target domain,

\begin{align}
\underset{F_s, F_t}{min} \:  \underset{D}{max} \: L(D, F_s, F_t) =   E_{x \sim P_{s}}[log D(F_s(X))] + E_{x \sim P_{t}}[log(1 - D(F_t(X))].
\label{eq1}
\end{align}

%\noindent where $D$ is the domain classifier, and $F_s$ is the feature extractor of source data and $F_t$ is the feature extractor for target data. The discriminator, $D$ classify the features of the source and target domains. 

%\begin{algorithm*}
  %\caption{Training procedure of our proposed method}\label{algorithom}
 % \begin{algorithmic}[1]
      %\For{\texttt{the $n$ number of training iterations}}
       % \State \texttt{$N_s$ number of source samples}
        
        %\State \texttt{$N_t$ number of target samples} \newline 
       % Calculate 

      %\EndFor
  %\end{algorithmic}
%\end{algorithm*}

%\begin{algorithm}
  %\caption{Mini-batch Training procedure of our proposed method}\label{algorithom}
 % \begin{algorithmic}[1]
      %\For{the $n$ number of training iterations}
        %\State{ $N_s$ number of source data with labels $\{X_i^s, Y_i^s $\}.}
        %\State { $N_t$ number of target data without labels $\{X_i^t$\}.}
        
       % \State{ Calculate losses following Equations \ref{eq3}, \ref{eq1}} and  \ref{eq4}. %Update the parameter of $\gamma$ and $\sigma$ following Equation \ref{eq8}.

 %\State{Update the parameter of $\gamma$ and $\sigma$ following Equation \ref{eq8}}
        
     % \EndFor
 % \end{algorithmic}
%\end{algorithm}

The network for feature extraction for source and target domains can use either an identical framework \cite{pmlr-v37-ganin15} where the weights are shared between the networks or a distinct framework \cite{8099799} where the weights are not shared. In this paper, we use the unshared feature extractors for the source and target domains to achieve more domain specific features. We follow a similar procedure as  \cite{8099799} where the classification loss is used with the source feature extractor as follows,

\begin{align}
\underset{F_s, C}{min} \: L_s =   E_{x \sim P_{s}} L_c(C(F_s(X_i^s)),Y_i^s),
\label{eq2}
\end{align}

\noindent where $L_c$ is the cross entropy loss for the source domain classification task as we have the supervised data for this domain. Cross entropy loss is defined as follows with a set of input source data $X_i^s$ = $\{ X_1^s, X_2^s,...,X_n^s\}$ with the corresponding labels $Y_i^s$ = $\{Y_1^s, Y_2^s,...,Y_n^s \} $ where $i = 1, 2, 3, ..., N$,

\begin{align}
\underset{X_i^s, Y_i^s}{min}\: L_c = - \sum_{i}^{K_{}} y_{i} \log (p_i),
\label{eq3}
\end{align}
where $K$ is the number of categories, $y_i$ is the ground truth and $p_i$ is the predicted value.
\subsection{Correlation-aware Adversarial Domain Adaptation (CAADA)}

The objective of DA is to minimize the discrepancy of the source and target data as much as possible during the training stage without utilizing the target labels due to unavailability. Figure \ref{fig:architecture} shows our proposed method for unsupervised domain adaptation based on adversarial learning along with a correlation alignment module during the training phase. The correlation alignment metric reduces the domain discrepancy using the second order statistics of the source and target data. We define the coral loss of the source and target activation features as \cite{dcoral},

\begin{equation}
\min_{F_s,F_t} L_{DM} (F_s, F_t) = \frac{1}{4d^2} \|C_s-C_t\|_F^2,
\label{eq4}
\end{equation}

\noindent where $C_s$ and $C_t$ denote the features covariance matrices of the source and target data, $d$ indicates the dimension of the activation features and $||.||_F^2$ denotes the squared matrix Frobenius norm. The $C_s$ and $C_t$ are given by the following equations \cite{dcoral},

\begin{equation}
C_s = \frac{1}{N_s-1} (F_s^TF_s-\frac{1}{N_s}(1^TF_s)^T(1^TF_s)) , 
\label{eq5}
\end{equation}

\begin{equation}
C_t = \frac{1}{N_t-1} (F_t^TF_t-\frac{1}{N_t}(1^TF_t)^T(1^TF_t)).
\label{eq6}
\end{equation}

After incorporating the correlation alignment module between the last fully connected layers of the source and target feature extraction networks to minimize the divergence of the source and target data again, the objective function of CADA is,

\begin{align}
\underset{F_s, F_t}{min} \:  \underset{D}{max} \: L(D, F_s, F_t) + L_{DM}(F_s, F_t) =  E_{x \sim P_{s}}[log D(F_s(X))] + \nonumber	\\ E_{x \sim P_{t}}[log(1 - D(F_t(X))] + \sigma L_{DM}(F_s, F_t),
\label{eq7}
\end{align}

\noindent where $\sigma$ is a hyper parameter that is fine-tuned. We discuss the procedure of fine tuning hyper parameters in Section \ref{hyper}.

The correlation alignment  metric forces the model to minimize the source and target domain discrepancy further which helps to obtain a more domain agnostic model that can be applied to classify unlabeled target data during the test phase more effectively. The overall objective of our proposed model is,

\begin{align}
\underset{X_s, Y_s}{min}\: L_c + \underset{F_s, F_t}{min} \:  \underset{D}{max} \: \gamma L(D, F_s, F_t) + \sigma L_{DM}(F_s, F_t),
\label{eq8}
\end{align}
where $\gamma$ and $\sigma$ are the hyper-parameters for adversarial and distribution matching loss that are needed to be fine tuned during training. 
%We update these parameters during training. 

%The algorithm is shown in Alg. \ref{algorithom}.

\subsection{Correlation-aware Adversarial Domain Generalization (CAADG)}

The difference between DA and DG approaches is that the target data is unavailable during the training stage in the latter. We need to fully utilize all available source domains. We use the same architecture as our DA model except we discard the target data in the training phase. For domain generalization, the training data always contains more than one source domain. Most of the existing domain generalization methods \cite{8658643,DBLP:conf/iccv/GhifaryKZB15,8053784,Li2018eccv} split the source data as $70 \%$ - $30\%$ and aggregate these $70 \%$ data as training data and $30\%$ data as validation data from all the source domains. Aggregating the data from the source domains and training a single deep neural network on all the data and testing on the unknown target domain provides a strong domain agnostic model that outperforms many prior approaches \cite{li2019episodic}. The discrepancy is minimised between these training and validation data during training. We also adopt the same setting for fair comparison. For domain generalization, the target data is always unknown. The main objective of domain generalization is to achieve a domain agnostic model from the available source domains and apply it to the unknown target data. Splitting and aggregating the source data also reduces the domain discrepancy. We split all the source data $70\% - 30\%$ as training and validation sets. We feed $70\%$ of the data data (aggregated from all source domains) in one stream of the CNN and $30\%$ data (aggregated from all source domains) into another stream of the CNN. Splitting and aggregating data this way, and feeding the data through two streams of the CNN is a simple yet effective method to minimize the discrepancy among domains. However, this approach does still suffer from some domain mismatch which is not removed completely. In Figure \ref{fig:Loss} below, we visualize the classification loss, the discrepancy loss and the discriminator loss. As the discrepancy loss is applied in the last fully connected layer and it is randomly initialized with N (0, 0.005), the discrepancy loss is very small while the classification and discriminator losses are large. From Figure \ref{fig:Loss}, we can see that after splitting and aggregating the source data, the domain mismatch still exists between the data which is feed through the two streams of the CNN. The adversarial learning module forces to reduce the discrepancy among the source domains in the $fc_{B}$ layer whereas the correlation alignment module forces the domains to be aligned in the last fully connected layer ($fc8$). Introducing the correlation alignment module with adversarial learning, the nework is more capable to reduce the domain shifts among the source domains and obtains a domain agnostic model that can be applied for the target data for a certain task.

\begin{figure}[htbp!]
\begin{center}
\includegraphics[width=1.0\linewidth]{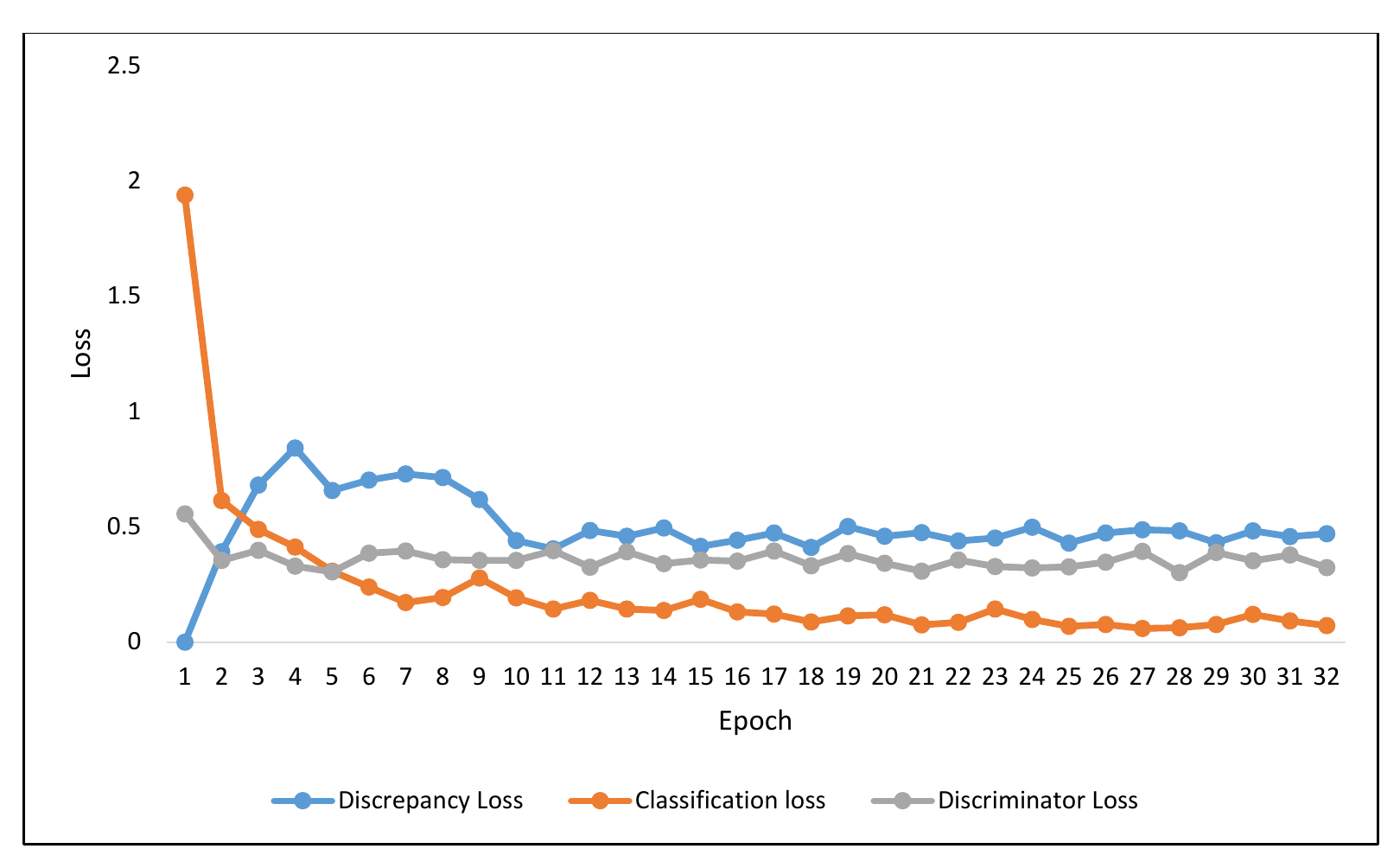}
\end{center}
   \caption{ Discrepancy loss, classification loss and discriminator loss for training the P, C, S $\rightarrow$ A transfer task on PACS dataset.}
\label{fig:Loss}
\end{figure}

The main difference between our work and prior adversarial domain adaptation methods \cite{pmlr-v37-ganin15, 8099799} is that our model forces the minimization of the source and target discrepancy twice. The domain discrepancy is minimized in $fc_{B}$ layer using adversarial learning, and  in $fc8$ layer using the correlation alignment metric. As a result, our model achieves the minimum discrepancy between the source and target data in domain adaptation. We also extend our model to make it suitable for DG scenarios where the target data is unavailable during the training phase and the discrepancy is minimized among the source domains. With training on multiple source domains' labeled data, the model can be applied to any unseen target domain without the adaptation step in the test phase. DG approaches are more suitable than DA methods in real-world scenarios since DG methods do not focus on the generalization capability on the particular target domain, but aim to generalize to any unseen target domain. DG considers the situation where the labeled data are collected from several source domains so that the trained model can handle new domains without the adaptation step in the future.

  %The main difference between our work and prior adversarial domain adaptation methods \cite{pmlr-v37-ganin15, 8099799} is that our model forces to minimize the source and target discrepancy twice using another distribution matching metric which is applied in between the last fully connected layers of the source and target feature extractor network. It is noted that when a distribution matching metric would be used with adversarial learning, the question would be arisen that where we can use distribution matching metric in a standard DNN. We find a way that we extend target feature extractor network by adding a fully connected layer as the source feature extractor network where the output vector is same as the number of total categories of a dataset. In between the last fully connected layers of the source and target feature extractor network, we use distribution matching metric. This metric forces the source and target distribution to be more closer in the last layer. In \cite{pmlr-v37-ganin15}, they used same feature extractor for the source and target data whereas we use two feature extractors for the source and target data respectively to get more domain specific features. They did not take account the last fully connected layer's features for the target data whereas we account the last fully connected layer features of both source and target data and we match these features during training using distribution discrepancy metric. 

%-------------------------------------------------------------------------

\section{Experiments}

In this section, we describe the experiments we have conducted to evaluate our method and compare the proposed method with state-of-the-art domain adaptation and generalization methods.

\subsection{Datasets}
The proposed approach is evaluated on five commonly used datasets in the context of image classification.
%\subsubsection*{Office-31}

\textbf{Office-31} \cite{Saenko:2010:AVC:1888089.1888106} dataset has three domains: Amazon (A), Webcam (W) and DSLR (D). The Amazon domain is formed with the downloaded images from amazon.com and this domain consists of 2817 images. The Webcam domain consists of the images that are captured by a webcam and it contains 795 images. The DSLR domain consists of images that are taken by a DSLR camera and it has 498 images. The number of categories of this dataset is 31. 

\textbf{Office-Home} \cite{venkateswara2017Deep} dataset has four domains: Clipart (C), Art (A), Real-world (R) and Product (P), and every domain consists of 65 categories. The Clipart domain is formed with clipart images. The Art domain consists of artistic images in the form of paintings, sketches, ornamentation etc. The Real-world domain's images are captured by a regular camera and the Product domain's images have no background. 

\begin{figure}
\begin{center}
\includegraphics[width=1.0\linewidth]{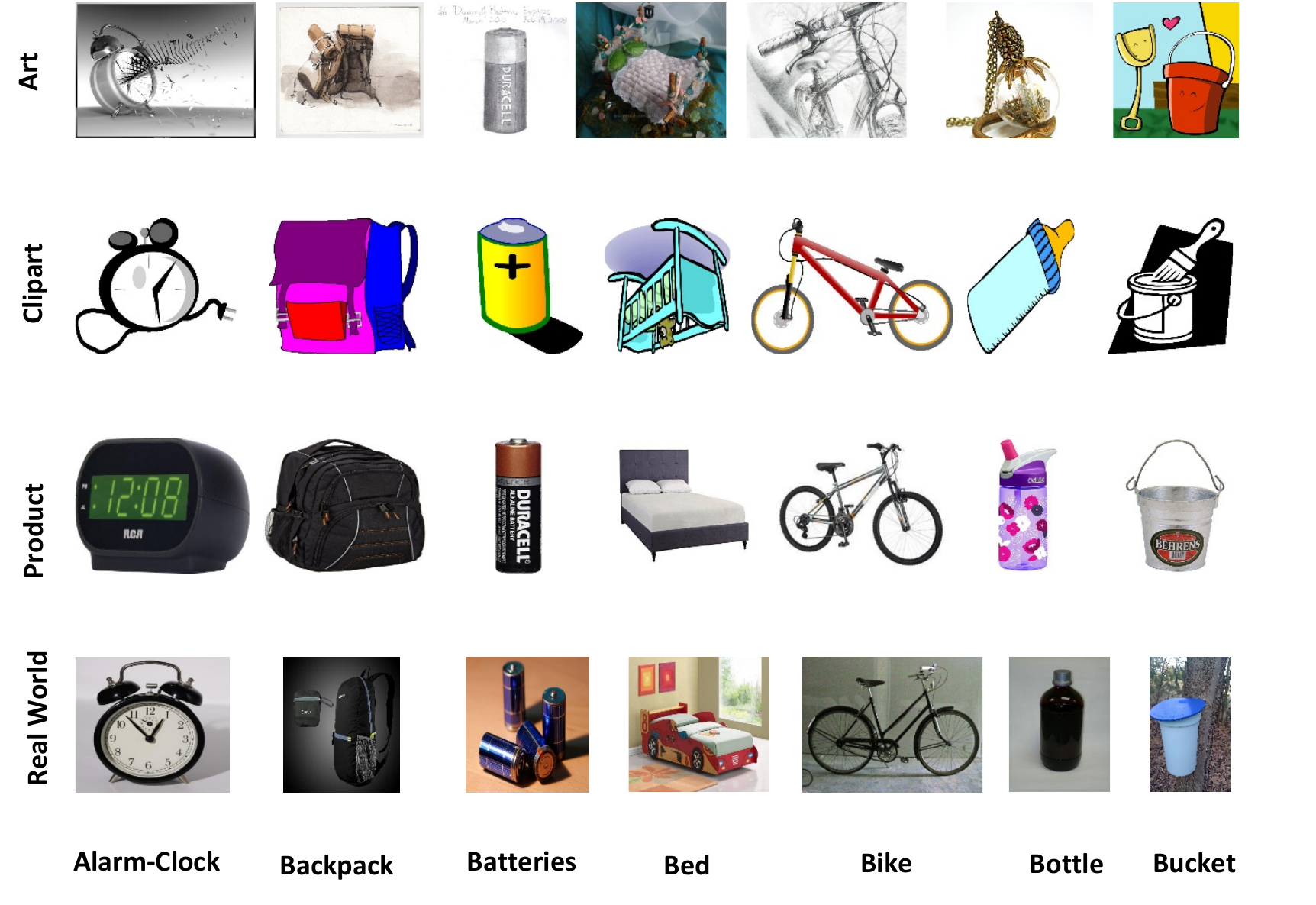}
\end{center}
   \caption{There are some example images that are taken from \textbf{Office-Home} dataset. It is a newly released and most challenging dataset in the domain adaptation community which consists of 65 categories.} It comprises of images of everyday objects. This dataset divided into 4 different domains; the \textbf{Art} domain consists of sketches, paintings, artistic images, the \textbf{Clipart} domain comprises clipart images, the \textbf{Product} domain comprises images which have no background and finally, the \textbf{Real-World} domain is created by taking images which are captured with a regular camera. The figure shows sample images from 7 of the 65 classes.
\label{fig:office}
\end{figure}

Figure \ref{fig:office} presents some sample images of 7 categories of Office-Home dataset.

%\subsubsection*{Office-Caltech}

%Office-Caltech is an another benchmark dataset for domain adaptation which is formed by taking the 10 common categories of two datasets: Office-31 and Caltech-256. It has 4 domains: Amazon (A), Webcam (W), DSLR (D) and Caltech (C). 

%\subsubsection*{ImageCLEF-DA}

\textbf{ImageCLEF-DA} is a benchmark dataset in the domain adaptation community for the ImageCLEF 2014 domain adaptation challenge. This dataset is formed with selecting 12 common categories shared by three popular datasets for object recognition: Caltech-256 (C), ImageNet ILSVRC 2012 (I) and Pascal
VOC 2012 (P), and each is considered as a domain. There are 600 images in each domain and 50 images in each category.

\textbf{ Office-Caltech} \cite{DBLP:conf/cvpr/GongSSG12} is another popular benchmark dataset in the domain adaptation community. It is formed by considering the 10 common categories of two datasets: Office-31 and Caltech-256. It has 4 domains: Amazon (A), Webcam (W), DSLR (D) and Caltech (C).

%We conduct experiments on 4 transfer tasks in the context of domain generalization where 3 domains or 2 domains are used as source domains and the target data is unseen in the training phase ($W, D, C$ $\rightarrow$ $A$; $A, W, D$ $\rightarrow$ $C$; $A, C$ $\rightarrow$ $D, W$ and $D, W$ $\rightarrow$ $A, C$).

%\subsubsection{PACS} 
\textbf{PACS} \cite{8237853} is a standard dataset for DG. It is formed by taking the common classes among Sketchy, Caltech256, TU-Berlin and Google Images. It has 4 domains and each domain consists of 7 clasess. It contains total 9991 images.

%\footnote{http://imageclef.org/2014/adaptation}

\subsection{Network Architecture}

In our method, we adopted the AlexNet \cite{NIPS2012_4824} architecture where 5 convolution ($conv1$, $conv2$, $conv3$, $conv4$ and $conv5$) layers and 3 fully connected layers ($fc6$, $fc7$ and $fc8$) are used for extracting features for the source and target data. A bottleneck layer ($fc_B$) with 256 units is added after $fc7$ layer as domain-adversarial training of neural
networks \cite{pmlr-v37-ganin15} so that safer transfer learning can be obtained. Due to dataset shift, the last-layer features are tailored to domain-specific structures that are not safely transferable, hence we add a bottleneck layer fcB. Gradient reversal layer ensures that the feature distributions over the two domains are made similar (as indistinguishable as possible for the domain classifier), thus resulting in the domain-invariant features. We use $fc_B$ as inputs to the discriminator and the $fc8$ layer for both streams of CNNs. The dimensions of the $fc8$ layer is set as the number of classes of the dataset (31 for Office-31 dataset). The adversarial module is used between the $fc_B$ layer of the source and target feature extraction network and the correlation alignment module is used between $fc8$ layers. We adopt the similar architecture as \cite{pmlr-v37-ganin15} for discriminator for a fair comparison. Our adversarial discriminator consists of 3 fully connected layers: two layers with 1024 hidden units followed by the final discriminator output. Each of the 1024-unit layers uses a ReLU activation function. We finetune the $conv1 - fc7$ layers with the pretrained AlexNet on ImageNet datasets \cite{imagenet_cvpr09}.

\subsection{Experimental Setup}

We use the Caffe \cite{jia2014caffe} framework to implement our proposed approach. We conduct all the experiments 3 times using high-performance computing (HPC) that consists of graphics processing unit (GPU), and we take the average accuracy in terms of image classification. For training the model, we set the batch size as 128 for all experiments. We set the learning rate to 0.001, momentum to 0.9 and weight decay to $5$ $\times$ $10^{-4}$ for optimizing the network. We follow the standard protocol for unsupervised domain adaptation where labeled source data and unlabeled target data are used for all transfer tasks as \cite{pmlr-v37-ganin15}. We also follow the standard protocol for domain generalization transfer tasks as \cite{Li2018eccv} where the target data is unavailable in the training phase.

%The classification accuracy of a model, $A_i$ depends on the images correctly identified. We evaluated all the methods by using the following formula:   

 %We follow the standard protocol for unsupervised domain adaptation technique where the source data are labeled and target data are unlabeled.
%We extend AlexNet architecture \cite{NIPS2012_4824} pretrained on Imagenet \cite{imagenet_cvpr09} dataset.

%\begin{equation}
%{A_i}  = \frac{t}{n} \times {100},
%\end{equation}

%\noindent where, t is the total number of correctly classified images, and n belong to the total images.

%\restylefloat{table}
\begin{table*}[!htbp]
\fontsize{7}{8}\selectfont 
%\small\addtolength{\tabcolsep}{0.0pt}
\begin{center}
 \resizebox{12cm}{!}{
\begin{tabular}{|l||c|c|c|c|c|c||c|c}

%\begin{tabular}{|p{1.5cm}|p{0.98cm}|p{0.98cm}|p{0.98cm}|p{0.98cm}|p{0.98cm}|p{0.98cm}||p{0.60cm}|}

%\rowcolor[gray]{0.95}
\hline

%\textbf{Methods} & \textbf{$A\rightarrow$W} &
\textbf{Methods} & \textbf{$A \rightarrow W$} & \textbf{$D\rightarrow W$} & \textbf{$D \rightarrow A$} & \textbf{$W \rightarrow A$}& \textbf{$W\rightarrow D$} & \textbf{$A \rightarrow D$} & \textbf{Avg.}\\
\hline\hline

%TCA  &21.5 &50.1  &8.0 &14.6 &58.4 &11.4 &27.3\\
%GFK &19.7 &49.7  &7.9  &15.8 &63.1 &10.6 &27.8\\

%VGG16  &63.9 &81.6 &46.9 &\textbf{54.1} &91.9 &63.1 &66.9\\
AlexNet \cite{NIPS2012_4824}  &61.6 &95.4 &51.1 &49.8 &99.0 &63.8 &70.1\\

DDC \cite{DBLP:journals/corr/TzengHZSD14} &61.8 &95.0 &52.1 &52.2 &98.5 &64.4 &70.7\\

DANN \cite{pmlr-v37-ganin15} &73.0 &96.4  &53.4  &51.2 &99.2 &72.3 &74.3\\

%&73.9 &94.9 &99.5 &73.8 &69.8  &53.8&77.6\\

D-CORAL \cite{dcoral}   &66.4 &95.7  &52.8  &51.5 &99.2 &66.8 &72.1\\

DAN \cite{DBLP:conf/icml/LongC0J15}  &68.5 &96.0  &54.0  &53.1 &99.0 &67.0  &72.9\\
%WMMD \cite{DBLP:journals/corr/YanDLWXZ17}  &66.8 &95.9 &98.7 &64.5 &53.8  &52.7&72.1\\%
%DDC \cite{DBLP:journals/corr/TzengHZSD14}  &61.5 &95.3 &98.5 &64.9 &47.2 &49.4&69.5\\%

DRCN \cite{DBLP:journals/corr/GhifaryKZBL16}  &68.7 &96.4  &56.0  &54.9 &99.0 &66.8  &73.6\\

RTN \cite{DBLP:conf/nips/LongZ0J16}   &73.3 &96.8  &50.5  &51.0 &99.6 &71.0   &73.7\\

JAN \cite{DBLP:conf/icml/LongZ0J17} &74.9 &96.6 &\textbf{58.3} &55.0 &99.5 &71.8 &76.9 \\

DAH \cite{venkateswara2017Deep}   &68.3 &96.1  &55.5  &53.0 &98.8 &66.5  &73.0\\

ADDA \cite{8099799} &73.5 &96.2 &54.6 &53.5 &98.8 &71.6 &74.7 \\

AutoDIAL \cite{carlucci2017auto} &75.5 &96.6 &58.1 &\textbf{59.4} &99.5 &73.6 &77.1 \\

MADA \cite{mada2018} &78.5 &\textbf{99.8} &56.0 &54.5 &\textbf{100.0} &74.1 &77.1 \\

%MSTN &80.5 &96.9 &62.5 &60.0 &99.9 &74.5 &79.1\\

\hline
\hline
\textbf{CAADA (Ours)}  & \textbf{80.2} &97.1  &58.1  &57.4 &99.2 &\textbf{77.7}  &\textbf{78.3}\\
\hline
\end{tabular}}
\end{center}
\caption{Image classification accuracies for deep domain adaptation on the Office-31 dataset. We use the conventional protocol for unsupervised domain adaptation where source data are labeled, but target data are unlabeled. $A \rightarrow W$ indicates $A$ (Amazon) is the source and $W$ (Webcam) is the target domain.}
\label{office31}
\end{table*}

%\restylefloat{table}
\begin{table*} [!htbp]
\fontsize{5}{7}\selectfont 
\begin{center}
%\begin{tabular}{|l||c|c|c|c|c|c|c|c|c|c|c|c||c|c}
%\rowcolor[gray]{0.95}

 \resizebox{12cm}{!}{
\begin{tabular}{|p{1.2cm}||p{0.40cm}|p{0.40cm}|p{0.40cm}|p{0.40cm}|p{0.40cm}|p{0.40cm}|p{0.40cm}|p{0.40cm}|p{0.40cm}|p{0.40cm}|p{0.40cm}|p{0.40cm}||p{0.40cm}|}

\hline

\textbf{Methods} & \textbf{$A\rightarrow C$} & \textbf{$A \rightarrow P$ } & \textbf{$A \rightarrow R$} & \textbf{$C \rightarrow A$} & \textbf{$C\rightarrow P$} & \textbf{$C\rightarrow R$}& \textbf{$P\rightarrow A$}& \textbf{$P  \rightarrow C$}& \textbf{$P \rightarrow R$}& \textbf{$R \rightarrow  A$}& \textbf{$R \rightarrow C$}& \textbf{$R \rightarrow P$}&  \textbf{Avg.}\\
\hline\hline

%TCA  &19.93 &32.08 &35.71 &19.00 &31.36 &31.74 &21.92 &23.64 &42.12 &30.74 &27.15 &48.68 &30.34\\
%GFK &21.60 &31.72 &38.83 &21.63 &34.94  &34.20&24.52 &25.73 &42.92 &32.88 &28.96 &50.89 &32.40\\

  %VGG16   &30.40 &\textbf{45.92} &\textbf{57.54} &35.40 &48.67  &50.75 &\textbf{35.77} &30.51 &60.20 &49.62 &34.54 &64.00  &45.28\\
AlexNet \cite{NIPS2012_4824}   &27.40 &34.53 &45.04 &32.40 &43.90  &46.72 &29.76 &32.94 &50.20 &40.74 &35.07 &55.99  &39.74\\
DANN \cite{pmlr-v37-ganin15}   &33.33 &42.96 &54.42 &32.26 &49.13  &49.76 &30.49&  38.14 &56.76 &44.71 &42.66 &64.65  &44.94\\

D-CORAL \cite{dcoral}  &32.18 &40.47 &54.45 &31.47 &45.8  &47.29 &30.03 &32.33 &55.27 &44.73 &42.75 &59.40  &42.79\\
DAN \cite{DBLP:conf/icml/LongC0J15}  &30.66 &42.17 &54.13 &32.83 &47.59  &49.78 &29.07 &34.05 &56.70 &43.58 &38.25&62.73  &43.46\\

RTN \cite{DBLP:conf/nips/LongZ0J16}  &31.23 &40.19 &54.56 &32.46 &46.60  &48.25 &28.20 &32.89 &56.38 &45.53 &44.74 &61.28  &43.53\\

JAN \cite{DBLP:conf/icml/LongZ0J17} &\textbf{35.5}&46.1&\textbf{57.7}&\textbf{36.4} &\textbf{53.3} &54.5 &33.4 &\textbf{40.3} &60.1 &45.9 &\textbf{47.4} &\textbf{67.9} &\textbf{48.2} \\
DAH \cite{venkateswara2017Deep}   &31.64 &40.75 &51.73 &34.69 &51.93  &52.79 &29.91 &39.63 &60.71 &44.99 &45.13 &62.54  &45.54\\

\hline
\hline
%\textbf{2 MMD} &32.05 &40.75 &51.73 &34.69 &51.93  &52.79 &29.91 &39.63 &60.71 &44.99 &45.13 &62.54  &45.54\\%
%\textbf{2 CORAL} &33.24 &42.05 &55.57 &33.12 &47.25  &49.68 &32.52 &34.85 &57.98 &47.89 &43.05 &63.29  &-\\%
\textbf{CAADA (Ours)}  &35.32 &\textbf{46.24} &56.58 &34.89 &51.79 &\textbf{60.01} &\textbf{34.94} & 39.97 &\textbf{60.20} &\textbf{47.84} &44.51 &\textbf{67.94} &\textbf{48.19}\\
\hline
\end{tabular}}
\end{center}
\caption{Image classification accuracies for deep domain adaptation on the Office-Home dataset. We use the conventional protocol for unsupervised domain adaptation where source data are labeled, but target data are unlabeled. $A \rightarrow C$ indicates $A$ (Art) is the source domain and $C$ (Clipart) is the target domain.}
\label{office-home}
\end{table*}

%\restylefloat{table}
\begin{table*}[!htbp]
\fontsize{7}{8}\selectfont 
\begin{center}
 \resizebox{12cm}{!}{
\begin{tabular}{|l||c|c|c|c|c|c||c|c}
%\rowcolor[gray]{0.95}
\hline

%\textbf{Methods} & \textbf{$A\rightarrow$W} &
\textbf{Methods} & \textbf{$I\rightarrow P$} & \textbf{$P \rightarrow I$} & \textbf{$I \rightarrow C$} & \textbf{$C \rightarrow I$}& \textbf{$C \rightarrow P$} & \textbf{$P \rightarrow C$} & \textbf{Avg.}\\
\hline\hline

AlexNet \cite{NIPS2012_4824}  &66.2 &70.0  &84.3 &71.3 &59.3 &84.5 &73.9\\

DAN \cite{DBLP:conf/icml/LongC0J15}  &67.3 &80.5  &87.7&76.0 &61.6 &88.4 &76.9\\

DANN \cite{pmlr-v37-ganin15}  &66.5 &81.8  &89.0&79.8 &63.5 &88.7 &78.2\\

RTN \cite{DBLP:conf/nips/LongZ0J16} &67.4 &81.3  &89.5&78.0 &62.0 &89.1 &77.9\\

JAN \cite{DBLP:conf/icml/LongZ0J17}  &67.2 &82.8  &91.3 &80.0 &63.5 &91.0 &79.3\\

MADA \cite{mada2018} & \textbf{68.3} &83.0 &91.0 &80.7 &63.8 &\textbf{92.2} &79.8 \\

%MSTN  &67.3 &82.8  &91.5 &81.7 &65.3 &91.2 &80.0\\

\hline
\hline
\textbf{CAADA (Ours)}  &67.8 &\textbf{84.5}  & \textbf{91.7}  & \textbf{81.3} & \textbf{63.9} &91.8  &\textbf{80.2}\\
\hline
\end{tabular}}
\end{center}
\caption{Image classification accuracies for deep domain adaptation on the ImageCLEF-DA dataset. We use the conventional protocol for unsupervised domain adaptation where source data are labeled, but target data are unlabeled. $I \rightarrow P$ indicates $I$ (ImageNet) is the source domain and $P$ (Pascal VOC) is the target domain.}
\label{ImageCLEF-DA}
\end{table*}

%\begin{figure}
%\begin{center}
%\includegraphics[width=1.0\linewidth]{A_W.pdf}
%\end{center}
   %\%caption{Test Accuracy for $A \rightarrow W$ %transfer task.}
%\label{fig:AW}
%\end{figure}

%\restylefloat{table}
\begin{table}[!htbp]
\fontsize{7}{8}\selectfont 
\begin{center}
\begin{tabular}{|l||c|c|c||c|}
%\rowcolor[gray]{0.95}
\hline

%\textbf{Methods} & \textbf{$A\rightarrow$W} &
\textbf{Methods} & \textbf{$A, W \rightarrow D$} & \textbf{$A, D \rightarrow W$} & \textbf{$D, W \rightarrow A$} &  \textbf{Avg.}\\
\hline\hline

Undo-Bias \cite{Khosla:2012:UDD:2402940.2402953}    &98.45 &93.38  &42.43   &78.08   \\

UML \cite{6751316}  &98.76 &93.76  &41.65   &78.05   \\

L-SVM \cite{DBLP:journals/jmlr/FanCHWL08}  &98.84  &93.98  &44.14   &78.99   \\

MTAE \cite{DBLP:conf/iccv/GhifaryKZB15}  &98.97  &94.21  &43.67   &78.95  \\

DSN \cite{Bousmalis:2016:DSN:3157096.3157135} &99.02  &94.45  &43.98   &79.15  \\

DGLRC  \cite{8053784} & \textbf{99.44}  &95.28  &45.36   &80.02   \\

\hline
\hline
\textbf{CAADG (Ours)}  &99.34 &\textbf{96.17}  & \textbf{47.49}  & \textbf{81.0}\\
\hline
\end{tabular}
\end{center}
\caption{Image classification accuracies for domain generalization on the Office31 dataset. We use the conventional protocol for domain generalization where the target data is totally unseen during training. $A, W \rightarrow D$ indicates $A, W$ are the source domains and $D$ is the target domain.}
\label{Off31_DG}
\end{table}

%\restylefloat{table}
\begin{table*}[!htbp]
\fontsize{7}{8}\selectfont 
\begin{center}
\begin{tabular}{|l||c|c|c|c||c|c||c|c}
%\rowcolor[gray]{0.95}
\hline

%\textbf{Methods} & \textbf{$A\rightarrow$W} &
\textbf{Methods} & \textbf{$W, D, C\rightarrow A$} & \textbf{$A, W, D \rightarrow C$} & \textbf{$A, C \rightarrow D, W$} & \textbf{$D, W \rightarrow A, C$}&  \textbf{Avg.}\\
\hline\hline

Undo-Bias \cite{Khosla:2012:UDD:2402940.2402953}   &90.98 &85.95 &80.49 &69.98 &81.85 \\

UML \cite{6751316}   &91.02  &84.59  &82.29   &79.54  &84.36 \\

L-SVM \cite{DBLP:journals/jmlr/FanCHWL08}   &91.87  &86.38  &84.59   &81.17  &86.00 \\

MTAE \cite{DBLP:conf/iccv/GhifaryKZB15} &93.13  &86.15  &85.35   &80.52  & 86.28\\

DSN \cite{Bousmalis:2016:DSN:3157096.3157135} &93.58  &86.71  &85.76   &81.22  &86.81 \\

DGLRC \cite{8053784} &94.21  &87.63  &86.32   &82.24  &87.60 \\

\hline
\hline
\textbf{CAADG (Ours)}  &\textbf{95.74} &\textbf{88.91}  & \textbf{86.79}  &82.11 &\textbf{88.39}\\
\hline
\end{tabular}
\end{center}
\caption{Image classification accuracies for domain generalization on the Office-Caltech dataset. We use the conventional protocol for domain generalization where the target data is totally unseen during training. $W, D, C \rightarrow A$ indicates $W, D, C$ are the source domains and $A$ is the target domain.}
\label{Off_Cal_DG}
\end{table*}

%\restylefloat{table}
\begin{table*}[!htbp]
\fontsize{7}{8}\selectfont 
\begin{center}
\begin{tabular}{|l||c|c|c|c||c|c||c|c}
%\rowcolor[gray]{0.95}
\hline

%\textbf{Methods} & \textbf{$A\rightarrow$W} &
\textbf{Methods} & \textbf{$P, C, S\rightarrow A$} & \textbf{$P, S, A \rightarrow C$} & \textbf{$S, A, C \rightarrow P$} & \textbf{$P, C, A \rightarrow S$}&  \textbf{Avg.}\\
\hline\hline

L-SVM \cite{DBLP:journals/jmlr/FanCHWL08}  &41.80 &52.30 &55.15 &47.87 &49.28 \\

KDA \cite{812} &47.66 &53.29 &59.04 &48.21 &52.05 \\

%UB &42.48 &48.93 &55.57 &46.30 & \\

DICA \cite{Muandet:2013:DGV:3042817.3042820}  &47.46 &57.00  &55.93 &40.70 &50.27\\

%LRE-SVM   &- &-  &- &- &- \\

MTAE \cite{DBLP:conf/iccv/GhifaryKZB15}  &45.95 &51.11  &58.44 &49.25 &51.19\\

DSN \cite{Bousmalis:2016:DSN:3157096.3157135}   &61.13 &66.54  &83.25 &58.58 &63.38 \\

DBADG \cite{8237853}   &62.86 &66.97  &\textbf{89.50} &57.51 &69.21 \\

CIDDG \cite{Li2018eccv} &62.70 &69.73 &78.65  &\textbf{64.45}&68.88 \\

\hline
\hline
\textbf{CAADG (Ours)}  &\textbf{65.52} &\textbf{69.90}  & 89.16  &63.37  &\textbf{71.98}\\
\hline
\end{tabular}
\end{center}
\caption{Image classification accuracies for domain generalization on the PACS dataset. We use the conventional protocol for domain generalization where the target data is totally unseen during training. $P, C, S \rightarrow A$ indicates $P, C, S$ are the source domains and $A$ is the target domain.}
\label{PACS_DG}
\end{table*}

\subsection{Baseline Methods}

For Office-31, Office-Home and ImageCLEF-DA datasets, we compare our proposed method on the DA scenario with Alexnet (without adaptation) \cite{imagenet_cvpr09} and the state-of-the-art domain adaptation methods: deep domain confusion (DDC) \cite{DBLP:journals/corr/TzengHZSD14}, domain-adversarial training of neural networks (DANN) \cite{pmlr-v37-ganin15}, deep-CORAL (D-CORAL) \cite{dcoral}, deep adaptation networks (DAN) \cite{DBLP:conf/icml/LongC0J15}, deep reconstruction classification network (DRCN) \cite{DBLP:journals/corr/GhifaryKZBL16}, residual transfer network (RTN) \cite{DBLP:conf/nips/LongZ0J16}, joint adaptation network (JAN) \cite{DBLP:conf/icml/LongZ0J17}, domain adaptive hashing (DAH) \cite{venkateswara2017Deep}, adversarial discriminative aomain adaptation (ADDA) \cite{8099799}, automatic domain alignment layers (AutoDIAL) \cite{carlucci2017auto}, and multi-adversarial domain adaptation (MADA) \cite{mada2018}. For the PACS dataset, Office-31 and Office-Caltech, we compare our method on the DG scenario with the state-of-the-art DG methods: learned - support vector machine (L-SVM) \cite{DBLP:journals/jmlr/FanCHWL08}, kernel fisher discriminant analysis (KDA) \cite{812}, domain-invariant component analysis (DICA) \cite{Muandet:2013:DGV:3042817.3042820}, multi-task auto-encoder (MTAE) \cite{DBLP:conf/iccv/GhifaryKZB15}, domain separation network (DSN) \cite{Bousmalis:2016:DSN:3157096.3157135}, deeper, broader and artier domain generalization (DBADG) \cite{8237853}, conditional invariant deep domain generalization (CIDDG) \cite{Li2018eccv}, undoing the damage of dataset bias (Undo-Bias) \cite{Khosla:2012:UDD:2402940.2402953}, unbiased metric learning (UML) \cite{6751316}, multi-task autoencoders (MTAE) \cite{DBLP:conf/iccv/GhifaryKZB15} and deep domain generalization with structured low-rank constraint (DGLRC) \cite{8053784}.

%We follow the standard protocol for unsupervised domain adaptation method \cite{pmlr-v37-ganin15} where we have labeled source and unlabeled target data. 

\subsection{Hyper-parameter tuning}
\label{hyper}

For any unsupervised domain adaptation and generalization method, we need to fine tune the hyper-parameters to achieve more efficient results. In our method, we fine tune two weight balance hyper-parameters: $\sigma$ and $\gamma$. We first apply different values for these two parameters on the experiments for the transfer task of $A \rightarrow W$ and we have found that  $\gamma$ = $\sigma$ = 0.1 provides the best results. Then we use this optimal value for $\gamma$ and $\sigma$ for all other transfer tasks in our method. 

%We set $\gamma$ = $\sigma$ = 0.1 in all of our experiments so that each domain discriminator have equal rights to minimize the domain discrepancy.
%-------------------------------------------------------------------------

%\begin{table}
%\begin{center}
%\begin{tabular}{|l|c|}
%\hline
%Method & Frobnability \\
%\hline\hline
%Theirs & Frumpy \\
%Yours & Frobbly \\
%Ours & Makes one's heart Frob\\
%\hline
%\end{tabular}
%\end{center}
%\caption{Results.   Ours is better.}
%\end{table}

%-------------------------------------------------------------------------
\subsection{Results and Discussion}

In this section, we discuss the results on benchmark datasets on both unsupervised domain adaptation and domain generalization settings.

\subsubsection{Results on domain adaptation scenarios }

%\subsection*{Office-31}
\textbf{Office-31} We conducted 6 transfer tasks in the context of domain adaptation: $A \rightarrow W$, $D \rightarrow W$, $D \rightarrow A$, $W \rightarrow A$, $W \rightarrow D$ and $A \rightarrow D$ and the results are reported in Table \ref{office31}. We also calculated the average accuracy. It is worth noting that our proposed approach achieves state-of-the-art results on two transfer tasks: $A \rightarrow W$ and $A \rightarrow D$ and the average accuracy on six different tasks is $78.3 \%$. It is also important to note that our method outperforms DANN on all transfer tasks and this improvement on transfer tasks provides the justification of integrating the correlation alignment metric in the last fully connected layer which helps to obtain a more domain agnostic model. 

From the results, we make some important observations: (1) All the deep domain adaptation methods outperform the standard deep learning methods and it can be revealed that deep neural networks without domain adaptation cannot eliminate the issue of domain shift; (2) DANN trains an extra domain classifier to enforce the minimization of discrepancy and this outperforms standard deep neural networks by about 4\%. This improvement also indicates the importance of adversarial learning for minimizing the discrepancy between the source and target data; (3) The use of a distribution matching metric (RTN, D-CORAL) also brings significant improvement over a standard deep learning method; and (4) Incorporating a correlation alignment module with adversarial learning improves the average accuracy by $5\%$ which indicates the capability of reducing domain discrepancy with our model is better than the baseline adversarial methods.

%For more justification, we visualize the transfer task $A \rightarrow W$ to compare our proposed method with the previous adversarial adaptation learning method (DANN). 

%\subsection*{Office-Home}
\textbf{Office-Home} We conducted 12 transfer tasks of 4 domains on Office-Home dataset in the context of domain adaptation. We report the results in  Table \ref{office-home}. We achieve the state-of-the-art performance on 6 transfer tasks including $A \rightarrow P$, $C \rightarrow R$, $P \rightarrow A$, $P \rightarrow R$, $R \rightarrow A$ and $R \rightarrow P$. It is worth noting that JAN obtains state-of-the-art performance on the other six transfer tasks as it reduces the domain shift in joint distributions of the network activations of multiple task-specific layers. In contrast, our proposed approach match the marginal distributions of features across domains. The prior best average accuracy was 48.2 \% achieved by JAN. The average accuracy achieved by our proposed method is 48.19 \% which outperforms the baseline method DANN by 3.25 \%. Our method beats other state-of-the-art methods except JAN which also obtains the same performance in terms of average accuracy. 

From the results of Office-Home dataset, we make two note-worthy observations: (1) All the deep domain adaptation methods obtain better performance than standard deep learning methods even when more categories are present in the dataset. It is noted that in Office-Home dataset, the number of total categories is 65; and (2) Our model is still capable to decrease the domain discrepancy even when more categories exist in the datasets.

%\subsection*{ImageCLEF-DA}

\textbf{ImageCLEF-DA} We conducted 6 transfer tasks: $I \rightarrow P $, $P \rightarrow I $, $I \rightarrow C $, $C \rightarrow I $, $C \rightarrow P $, and $P \rightarrow C $. We report the results in Table \ref{ImageCLEF-DA}. In this dataset, every category has the same number of images thus the images are balanced, but each domain has only 600 images that could not be enough for training the network. Our method outperforms existing approaches in four transfer tasks: $P \rightarrow I $, $I \rightarrow C $, $C \rightarrow I $, and $C \rightarrow P $. The prior best average accuracy on this dataset was 79.8\% achieved by MADA, and our method achieved 80.2 \% average accuracy which sets a new state-of-the-art performance.

From the results of ImageCLEF-DA dataset, we make two important observations: (1) Deep domain methods still achieve better performance over standard deep learning methods when few categories are shared by the source and target data; and (2) Our model is still capable to decrease the domain discrepancy even when few categories are shared by the source and target domains.

\subsubsection{Results on domain generalization scenarios}

\textbf{Office-31} In the context of domain generalization, we conducted 3 transfer tasks: $A, W \rightarrow D$; $A, D \rightarrow W$ and $D, W \rightarrow A$ where the target data is unavailable during the training phase. We report the results in Table \ref{Off31_DG}. From the results we cam observe that our proposed domain method achieves state-of-the performance on two transfer tasks: $A, D \rightarrow W $ and $D, W \rightarrow A $ and the average accuracy on three transfer tasks is $81.0\%$.

\textbf{Office-Caltech} We conducted four transfer tasks: $W, D, C \rightarrow A$; $A, W, D \rightarrow C$; $A, C \rightarrow D, W$ and $D, W \rightarrow A, C$ on domain generalization settings.We report the results on Table \ref{Off_Cal_DG}. From the results we can observe that we achieved state-of-the-art performance on three transfer tasks and we also achieved the best average accuracy on four transfer tasks.

\textbf{PACS} We conducted 4 transfer tasks: $P,C,S \rightarrow A $; $P,S,A \rightarrow C $; $S,A,C \rightarrow P $ and $P,C,A \rightarrow S $ on DG scenario where the target data is unavailable during the training phase. We report the results in Table \ref{PACS_DG}. From the results we can see that our proposed approach obtains state-of-the-art performance on two transfer tasks: $P,C,S \rightarrow A $ and $P,S,A \rightarrow C $ and the average accuracy on four transfer tasks is $71.98\%$.

%From the results, we make one important observation that our model is applicable both on DA and DG scenarios.

From the results (Tables \ref{office31} to \ref{PACS_DG}), we note that the proposed architecture performs well on both domain adaptation and domain generalization settings.

\subsection{Visualization}

\begin{figure}
\begin{center}
\includegraphics[width=1.0\linewidth]{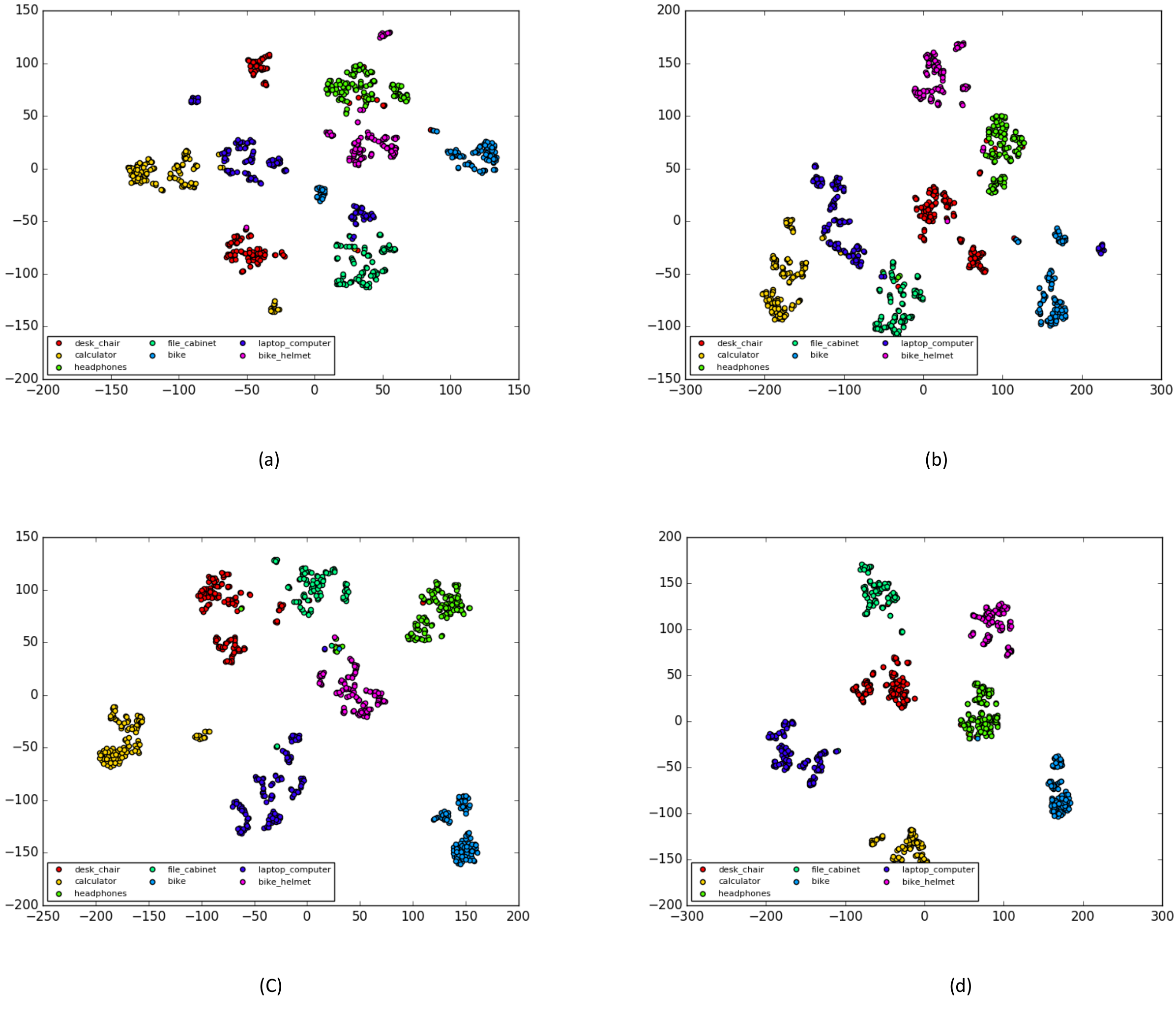}
\end{center}
   \caption{The t-SNE visualization of the activations of (a) AlexNet (without adaptation) (b) without adversarial loss (c) without discrepancy loss and (d) CAADA (ours).}
\label{fig:AW_CAFFENET}
\end{figure}

\begin{figure}
\begin{center}
\includegraphics[width=1.0\linewidth]{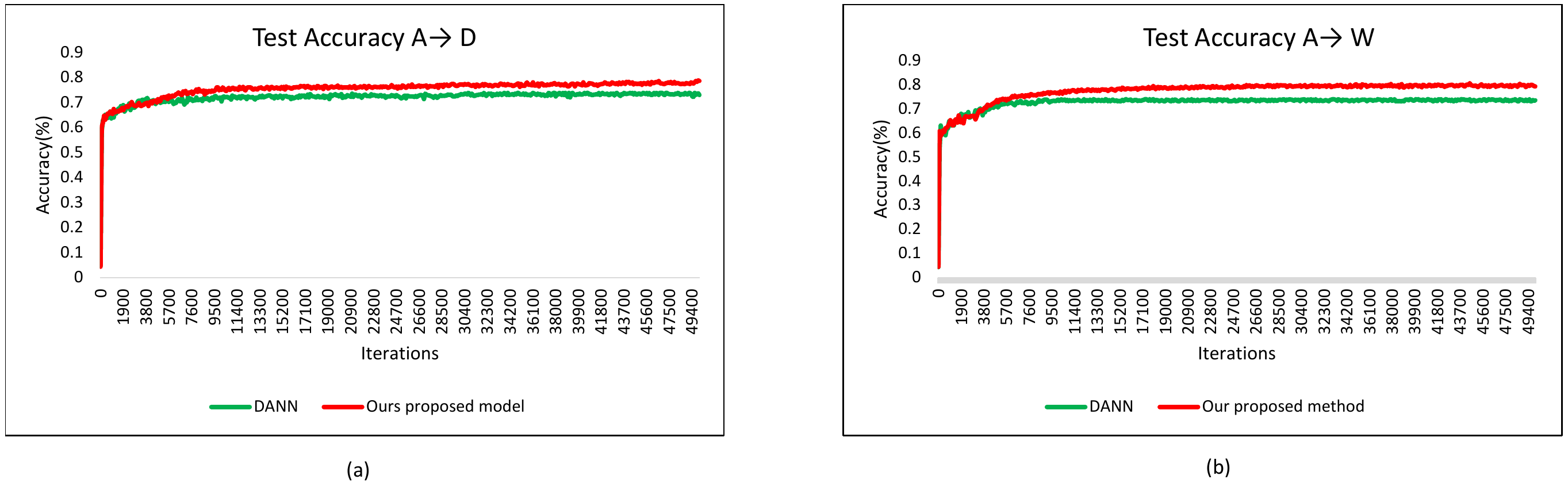}
\end{center}
   \caption{ Convergence: (a) Test Accuracy for $A \rightarrow D$ transfer task and (b) Test Accuracy for $A \rightarrow W$ transfer task.}
\label{fig:graph}
\end{figure}

We use t-SNE \cite{vanDerMaaten2008} to conduct an embedding visualization. We take images from Amazon and Webcam domains of Office-31 dataset to produce an embedding.
In Figure \ref{fig:AW_CAFFENET}, the representations in Amazon $\rightarrow$ Webcam transfer task is visualized. Figure \ref{fig:AW_CAFFENET}(a) shows the representation using standard deep learning (AlexNet) without any adaptation method. For better view, we have taken 7 different categories from Office-31 dataset. Figure \ref{fig:AW_CAFFENET}(b) shows the representation of our model without adversarial loss. Figure \ref{fig:AW_CAFFENET}(c) depicts the representations that are learned by our model withoout discrepancy loss. On the other hand, Figure \ref{fig:AW_CAFFENET}(d) depicts the representations that are learned by our proposed approach. 
Examining the embeddings, we find that the clusters created by our model separate the categories while mixing the domains much more effectively than standard DNN and using only discrepancy loss or only adversarial loss. These visualization also suggests that our proposed method is more capable to align the source and target features from the same class nearby.

\begin{figure}
\begin{center}
\includegraphics[width=1.0\linewidth]{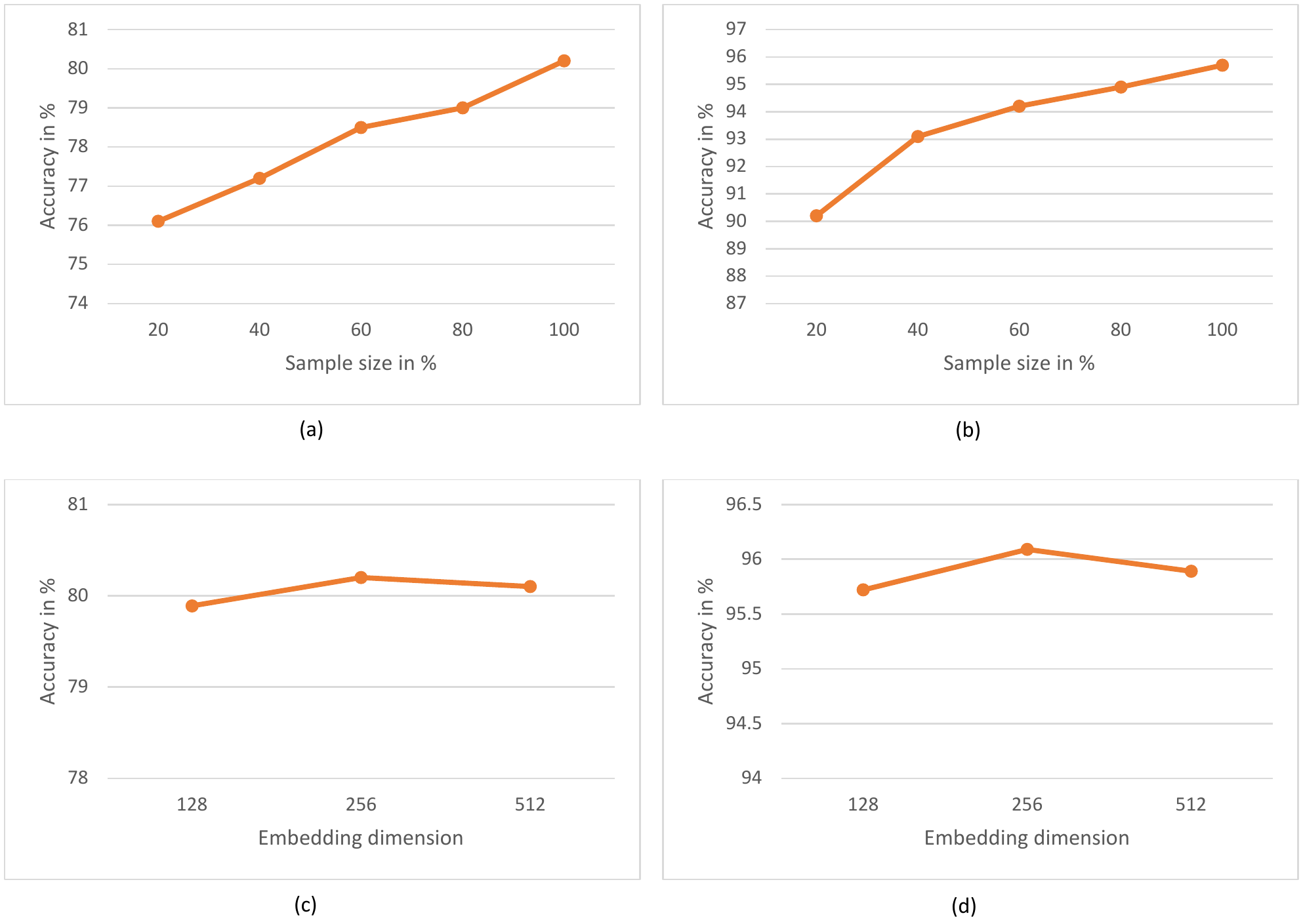}
\end{center}
   \caption{The accuracy with regard to various unlabeled target sample size and by varying feature embedding dimension: (a) sample size on $A \rightarrow W$ domain adaptation transfer task, (b) sample size on $W,D,C \rightarrow A$ domain generalization transfer task, (c) embedding dimension on $A \rightarrow W$ domain adaptation transfer task and (d) embedding dimension on $W,D,C \rightarrow A$ domain generalization transfer task.}
\label{fig:samplesize}
\end{figure}

\subsection{Convergence Performance}

Since our proposed method is based on adversarial learning, we compare the performance on convergence. The convergence is measured based on the test accuracy. Figure \ref{fig:graph}(a) and Figure \ref{fig:graph}(b) represents the test accuracy of our method and DANN \cite{pmlr-v37-ganin15} on $A \rightarrow D $ and $A \rightarrow W $ transfer tasks. The x-axis and y-axis of these accuracy graphs represent the iterations and accuracies in percentage. We have calculated the test accuracy over 50,000 iterations. From Figure \ref{fig:graph}, we can see that our model has the analogous convergence speed as DANN with remarkably better accuracy in the entire procedure of convergence.

\subsection{Sample Size of Target Domain}

In this section, we empirically show that our proposed method is data-driven. If the number of unlabeled target data is increased, the image classification performance is also boosted. We shuffle the target data on Office-31 for domain adaptation and Office-Caltech for domain generalization, and access the top $20\%$, $40\%$, $60\%$, $80\%$, $100\%$ data of each classes of the target domain. We train and evaluate our approach for both domain adaptation and domain generalization on $A$ $\rightarrow$ $W$ and $W, D, C$ $\rightarrow$ $A$ tasks respectively. As illustrated in Figure \ref{fig:samplesize}(a) and \ref{fig:samplesize}(b) , as the target data size moderately expands, the image recognition accuracy of the corresponding task increases accordingly. It demonstrates that when more unlabeled target data are involved in the training, a more transferable classifier with regard
to the target domain can be obtained.

\subsection{Sensitivity of Embedding Dimension}

In this part, we investigate the sensitivity of the embedding dimension of the bottleneck layer as it plays a significant role in reducing the discrepancy among domains. We conduct the experiments on  Office-31 and Office-Caltech datasets for domain adaptation and domain generalization tasks respectively.
We choose $A$ $\rightarrow$ $W$ transfer task for Office-31 on domain adaptation and $W, D, C$ $\rightarrow$ $A$ transfer task for Office-Caltech on domain generalization. We report the mean accuracy
with standard deviation for embedding dimensions varied
in \{128, 256, 512 \} respectively. As illustrated
in Figure \ref{fig:samplesize}(c) and \ref{fig:samplesize}(d), the accuracy almost maintains at the same level
and achieves slightly higher accuracy when the embedding dimension
is set to 256, implying that our approach is not sensitive
to the chosen range of feature space dimension.

\subsection{Complementary with correlation alignment}

In this part, we demonstrate that our approach is capable of cooperating
with correlation alignment. We conduct this case study on
Office-31 dataset on both domain adaptation and domain generalization tasks, and report the accuracy in Table \ref{Ablation study_office31_DA} and Table \ref{ablation_Off31_DG} respectively. As illustrated, with
correlation alignment to further reduce the discrepancy among the domains, we further boost the recognition performance and obtain $2.1 \%$ and $1.89 \%$ gain for domain adaptation and domain generalization tasks respectively.

%-------------------------------------------------------------------------

\subsection{Ablation Study}

Recent studies on domain adaptation suggested that both adversarial learning \cite{pmlr-v37-ganin15,8099799} and correlation alignment \cite{DBLP:conf/icml/LongC0J15,DBLP:journals/corr/ZellingerGLNS17,dcoral} can play a vital role to reduce the domain discrepancy due to dataset bias. It is noted that adversarial learning uses a domain classifier to reduce the discrepancy whereas a correlation alignment module calculates co-variances of the data and minimizes the distance among the data of different domains. But these methods cannot eliminate the discrepancy completely. Hence, we combined two different techniques to minimize the discrepancy so that we can estimate the minimum discrepancy of the available domains for both domain adaptation and domain generalization tasks.

%\restylefloat{table}
\begin{table*}[!htbp]
\fontsize{5}{6}\selectfont 
\begin{center}
 \resizebox{12cm}{!}{
\begin{tabular}{|l||c|c|c|c|c|c|c|c}

%\begin{tabular}{|p{1.5cm}|p{0.98cm}|p{0.98cm}|p{0.98cm}|p{0.98cm}|p{0.98cm}|p{0.98cm}||p{0.60cm}|}

%\rowcolor[gray]{0.95}
\hline

%\textbf{Methods} & \textbf{$A\rightarrow$W} &
\textbf{Methods} & \textbf{$A \rightarrow W$} & \textbf{$D\rightarrow W$} & \textbf{$D \rightarrow A$} & \textbf{$W \rightarrow A$}& \textbf{$W\rightarrow D$} & \textbf{$A \rightarrow D$} \\
\hline\hline

Correlation-alignment    &67.1 &94.9  &53.5 &52.3 &99.3 &67.3 \\

Adversarial-learning  &74.1 &97.0  &54.0  &52.7 &99.1 &73.1 \\

CAADA (Without Correlation alignment) &74.5 &97.2 &54.3  &53.0 &99.1 &75.6 \\

\hline
\hline
CAADA (With Correlation alignment)  & \textbf{80.2} &\textbf{97.3}  &\textbf{58.1}  &\textbf{57.4} &\textbf{99.2} &\textbf{77.7} \\
\hline
\end{tabular}}
\end{center}
\caption{Ablation study on Office-31 dataset in the context of domain adaptation.}
\label{Ablation study_office31_DA}
\end{table*}

%\restylefloat{table}
\begin{table}[!htbp]
\fontsize{5}{6}\selectfont 
\begin{center}
 \resizebox{12cm}{!}{
\begin{tabular}{|l||c|c|c|c|}
%\rowcolor[gray]{0.95}
\hline

%\textbf{Methods} & \textbf{$A\rightarrow$W} &
\textbf{Methods} & \textbf{$A, W \rightarrow D$} & \textbf{$A, D \rightarrow W$} & \textbf{$D, W \rightarrow A$} \\
\hline\hline
Correlation-alignment &95.25 &93.15 &45.29 \\
Adversarial-learning &97.34 &94.10 &45.35 \\
CAADG (Without Correlation alignment)  &97.51 &94.32  & 45.60 \\
\hline
\hline
CAADG (With Correlation alignment)  &\textbf{99.34} &\textbf{96.17}  & \textbf{47.49}  \\
\hline
\end{tabular}}
\end{center}
\caption{Ablation study on Office-31 dataset in the context of domain generalization.}
\label{ablation_Off31_DG}
\end{table}

In this section, we perform ablation study on Office-31 dataset in the context of both domain adaptation and domain generalization with different components ablation, i.e., training with only adversarial learning, training with only correlation alignment. Those experiments shows different contributions of components and provide more justification of adopting correlation alignment with adversarial learning. We compare the results with our method and the results of domain adaptation is reported in Table \ref{Ablation study_office31_DA}. We report the domain generalization performance on Office-31 dataset in Table \ref{ablation_Off31_DG}. 

From the results of ablation studay, it can be observed that incorporating correlation alignment with adversarial learning improves the discrimination and generalization ability. We also observe that in each transfer task of the domain adaptation and generalization, we get better performance than using only correlation alignment or using only adversarial learning to minimize the discrepancy between the domains.

%_____-------New

%\restylefloat{table}
\begin{table*}[!htbp]
\fontsize{5}{6}\selectfont 
\begin{center}
 \resizebox{12cm}{!}{
\begin{tabular}{|l||c|c|c|c|c|c||c|c}

%\begin{tabular}{|p{1.5cm}|p{0.98cm}|p{0.98cm}|p{0.98cm}|p{0.98cm}|p{0.98cm}|p{0.98cm}||p{0.60cm}|}

%\rowcolor[gray]{0.95}
\hline

%\textbf{Methods} & \textbf{$A\rightarrow$W} &
\textbf{Methods} & \textbf{$A \rightarrow W$} & \textbf{$D\rightarrow W$} & \textbf{$D \rightarrow A$} & \textbf{$W \rightarrow A$}& \textbf{$W\rightarrow D$} & \textbf{$A \rightarrow D$} & \textbf{Avg.}\\
\hline\hline

CAADA (2 Discriminator)  &73.3 &96.7 &54.1 &52.0 &99.3 &72.6 &74.67 \\

%MSTN &80.5 &96.9 &62.5 &60.0 &99.9 &74.5 &79.1\\

CAADA (without pre-trained)  & 80.2 &97.1  &58.1  &57.4 &99.2 &77.7  &78.3\\

CAADA (pre-trained on source data)  & \textbf{80.4} &\textbf{97.3}  &\textbf{59.2}  &\textbf{58.1} &\textbf{99.3} &\textbf{78.0}  &\textbf{78.7}\\
\hline

\end{tabular}}
\end{center}
\caption{Ablation study to show the effect of using two discriminator and pre-trained source model on Office-31 dataset in the context of domain adaptation.}
\label{ablation9office31}
\end{table*}

%\restylefloat{table}
\begin{table*}[!htbp]
\fontsize{5}{6}\selectfont 
\begin{center}
 \resizebox{12cm}{!}{
\begin{tabular}{|l||c|c|c|c|c|c||c|c}
%\rowcolor[gray]{0.95}
\hline

%\textbf{Methods} & \textbf{$A\rightarrow$W} &
\textbf{Methods} & \textbf{$I\rightarrow P$} & \textbf{$P \rightarrow I$} & \textbf{$I \rightarrow C$} & \textbf{$C \rightarrow I$}& \textbf{$C \rightarrow P$} & \textbf{$P \rightarrow C$} & \textbf{Avg.}\\
\hline\hline

CAADA (2 Discriminator)   & 66.3 &82.1 &89.6 &79.6 &63.9 &89.3 &78.47 \\

%MSTN  &67.3 &82.8  &91.5 &81.7 &65.3 &91.2 &80.0\\

CAADA (without pre-trained)  &67.8 &84.5  & 91.7  & 81.3 & 63.9 &91.8  &80.2\\

CAADA (pre-trained on source data) &\textbf{68.2} &\textbf{84.9}  & \textbf{92.4}  & \textbf{81.4} & \textbf{64.7} &\textbf{92.6}  &\textbf{80.7}\\
\hline

\end{tabular}}
\end{center}
\caption{Ablation study to show the effect of using two discriminator and pre-trained source model on ImageCLEF-DA dataset in the context of domain adaptation.}
\label{ablation10ImageCLEF-DA}
\end{table*}

%\restylefloat{table}
\begin{table*}[!htbp]
\fontsize{5}{6}\selectfont 
\begin{center}
 \resizebox{12cm}{!}{
\begin{tabular}{|l||c|c|c|c||c|c||c|c}
%\rowcolor[gray]{0.95}
\hline

%\textbf{Methods} & \textbf{$A\rightarrow$W} &
\textbf{Methods} & \textbf{$P, C, S\rightarrow A$} & \textbf{$P, S, A \rightarrow C$} & \textbf{$S, A, C \rightarrow P$} & \textbf{$P, C, A \rightarrow S$}&  \textbf{Avg.}\\
\hline\hline

CAADG (2 Discriminator)  &63.42 &67.28 &86.39  &61.45 &69.64 \\

CAADG (without pre-trained)   &65.52 &69.90  & 89.16  &63.37  &71.98\\

CAADG (pre-trained on source data)  &\textbf{66.15} &\textbf{70.29}  & \textbf{89.68}  &\textbf{63.74}  &\textbf{72.47}\\
\hline

\end{tabular}}
\end{center}
\caption{Ablation study to show the effect of using two discriminator and pre-trained source model on PACS dataset in the context of domain generalization.}
\label{ablation11PACS_DG}
\end{table*}

%\restylefloat{table}
\begin{table*}[!htbp]
\fontsize{5}{6}\selectfont 
\begin{center}
 \resizebox{12cm}{!}{
\begin{tabular}{|l||c|c|c|c||c|c||c|}
%\rowcolor[gray]{0.95}
\hline

%\textbf{Methods} & \textbf{$A\rightarrow$W} &
\textbf{Methods} & \textbf{$W, D, C\rightarrow A$} & \textbf{$A, W, D \rightarrow C$} & \textbf{$A, C \rightarrow D, W$} & \textbf{$D, W \rightarrow A, C$}&  \textbf{Avg.}\\
\hline\hline

CAADG (2 Discriminator) &93.81  &85.98  &86.09   &81.36  &86.81 \\

CAADG (without pre-trained)   &95.74 &88.91  & 86.79  &82.11 &88.39\\

CAADG (pre-trained on source data)  &\textbf{96.09} &\textbf{89.12}  & \textbf{87.35}  &\textbf{82.67} &\textbf{88.81}\\
\hline

\end{tabular}}
\end{center}
\caption{Ablation study to show the effect of using two discriminator and pre-trained source model on Office-Caltech dataset in the context of domain generalization .}
\label{ablation12Off_Cal_DG}
\end{table*}

We examined the performance of incorporating another discriminator to align the features of Fc8. The results are reported in Tables \ref{ablation9office31} - \ref{ablation12Off_Cal_DG}. According to the results, we can observe that even using another discriminator to align  features of Fc8  instead of using the correlation alignment loss still suffer from poor performance than our proposed method. In ADDA \cite{8099799}, the researchers use the pre-trained source model as an initialization. On the other hand, other adversarial learning methods JAN \cite{DBLP:conf/icml/LongZ0J17}, DANN \cite{pmlr-v37-ganin15}, RTN \cite{DBLP:conf/nips/LongZ0J16} did not use a pre-trained source model. We adopted the approach from JAN, DANN and RTN. We examined the approach of using the pre-trained source model as an initialization. From the results (Tables \ref{ablation9office31} - \ref{ablation12Off_Cal_DG}), we have seen that the improvements is about $0.4 \%$ to $0.5\%$.

\subsection{Discussion}

From the above analysis, it can be noted that all the deep domain adaptation and generalization methods outperform the standard deep learning approaches. Adversarial learning with correlation alignment improves the generalization capability of the model. When the target data size is increased, the image classification accuracy is also increased accordingly. It can be revealed that when more unlabeled target data are involved in the training, a more transferable classifier with regard to the target domain can be obtained. The image recognition accuracy almost maintains at the same level with varying the embedding dimension and achieves a slightly higher accuracy when the embedding dimension is set to 256. Using a pre-trained source model as an initialization improves the image recognition accuracy by $0.4 \%$  to $0.5 \%$ on domain adaptation and generalization tasks respectively.

%\subsubsection{Effect of different components}

%-------------------------------------------------------------------------

\section{Conclusion}

In this paper, we proposed a novel correlation-aware adversarial domain adaptation and generalization method which aims to minimize the domain discrepancy as much as possible during training stages using a correlation alignment metric along with adversarial learning. We have shown that our model can be applied to address the DG issue without changing the basic network architecture. We have found that using the correlation alignment along with adversarial learning for domain adaptation and domain generalization scenarios works efficiently. Experiments on different datasets on domain adaptation and domain generalization verify the effectiveness of our proposed method.

\section*{References}

\bibliography{mybibfile}

\end{document}